# Kelime Gömme Teknikleri ile Kitap Özetlerinin Kategorik Sınıflandırılması

Kerem KESKİN[1]*, Mümine KAYA KELEŞ[2]

[1]*Çukurova Üniversitesi, Mühendislik Fakültesi, Bilgisayar Mühendisliği, Adana, TÜRKİYE*
[2]*Adana Alparslan Türkeş Bilim ve Teknoloji Üniversitesi, Bilgisayar ve Bilişim Fakültesi, Bilgisayar Mühendisliği, Adana, TÜRKİYE*

**Özet**

Bu çalışmada kitap sitelerinden çekilen kitap özetleri ve kategorilerinin kelime gömme yöntemleri, doğal dil işleme teknikleri ve makine öğrenmesi algoritmaları ile sınıflandırılması yapılmıştır. Ayrıca sıklıkla kullanılan kelime gömme yöntemlerinden One Hot Encoding, Word2Vec ve Terim Frekansı - Ters Doküman Frekansı (TF-IDF) yöntemleri bu çalışmada kullanılmış ve başarı karşılaştırılması yapılmıştır. Ek olarak kullanılan ön işleme yöntemlerinin kombinasyon tablosu gösterilmiş ve başarı tablosuna eklenmiştir. Sonuçlara bakıldığında Türkçe metinler için Destek Vektör Makinesi, Naive Bayes ile Lojistik Regresyon Modellerinin ve TF-IDF ile One-Hot Encoder kelime gömme tekniklerinin daha başarılı sonuçlar verdiği gözlenmiştir.

*Anahtar Kelimeler: Veri Bilimi, Makine Öğrenmesi, Doğal Dil İşleme, Kelime Gömme*

## Categorical Classification of Book Summaries Using Word Embedding Techniques

**Abstract**

In this study, book summaries and categories taken from book sites were classified using word embedding methods, natural language processing techniques and machine learning algorithms. In addition, one hot encoding, Word2Vec and Term Frequency - Inverse Document Frequency (TF-IDF) methods, which are frequently used word embedding methods were used in this study and their success was compared. Additionally, the combination table of the pre-processing methods used is shown and added to the table. Looking at the results, it was observed that Support Vector Machine, Naive Bayes and Logistic Regression Models and TF-IDF and One-Hot Encoder word embedding techniques gave more successful results for Turkish texts.

*Keywords: Data Science, Machine Learning, Natural Language Processing, Word Embedding*

## 1 Giriş

Sağlıktan eğitime birçok sektörde makine öğrenmesi teknikleri ve doğal dil işleme teknikleri kullanılmaya devam edilmektedir. Birçok dilde doğal dil işleme çalışmaları sürmektedir ve türkçe dil yapısının zorluklarının aşılması için farklı tekniklere başvurulmaktadır. Bu tekniklerden ön işleme metotları ve kelime gömme algoritmaları sıklıkla çalışmalarda kullanılmaktadır. Yapılan bir çalışmada Word2Vec ve Terim Frekansı - Tersine Doküman Frekansı tekniklerinin kullanılması ile türkçe metinlerin sınıflandırılmasında başarıyı arttırdığı gözlenmiştir [1]. Ayrıca yapılan başka bir çalışmada Türkçe ve İngilizce Twitter mesajlarının duygu analizinde başarıyı arttırmak için Word2Vec kelime gömme tekniği kullanılmıştır [2]. Kelime gömme yöntemleri tek başına veya farklı yöntemlerle kombine edilerek de kullanılabilir ve sınıflandırma başarısını arttırabilir. Kelime gömme yöntemlerinin kombine edilerek kullanılması ile yapılan bir çalışmada Word2Vec ve TF-IDF kelime gömme teknikleri birleştirilmiş ve sınıflandırma başarısı için olumlu sonuçlar vermiştir [3]. Metinlerde kelimeler arasındaki anlam ilişkilerinin öğrenilmesi ve gruplandırılması için Word2Vec gibi kelime gömme yöntemlerine başvurulmaktadır. Otel yorumlarının duygu analizi ve sınıflandırılması için yapılan bir çalışmada başarıyı arttırmak için Word2Vec yöntemi kullanılmıştır ve kelime grupları arasındaki anlam

---

*Contact email: kkeskin@cu.edu.tr

ilişkileri gösterilerek pozitif ve negatif olarak sınıflandırılmıştır [4].

Türkçe dilinin yapısından dolayı oluşan problemleri çözmek için yapılan bu çalışmada türkçe kitap özetlerinin sınıflandırılması farklı kelime gömme teknikleri ve ön işleme metotları kullanılarak yapılmıştır ve başarı karşılaştırmaları tablo halinde gösterilmiştir. Yapılan birçok çalışmada ön işleme yöntemleri denenmektedir ama farklı kombinasyonları gösterilmemektedir. Bu çalışmada yapılan ön işlemelerin kombinasyonları her bir kelime gömme yöntemi için uygulanmış ve her bir kombinasyon için en yüksek başarıyı veren modelin f-skoru ve başarı oranı tabloda gösterilmiştir.

Bu çalışmada, algoritmaların örüntüleri ayırt etme, temaları belirleme ve özetlerin içeriği hakkında tahminler yapma konusunda güçlendirilmesi amaçlanmaktadır. Kelime gömme yöntemleri ve makine öğreniminin bu sentezi, yalnızca kategorizasyon sürecini kolaylaştırmakla kalmıyor, aynı zamanda Türkçe dilinin inceliklerine dair daha derin içgörüler arayan araştırmacılar ve akademisyenler için dinamik bir araç sunuyor. Bu çalışmada gelişmiş Doğal dil işleme metodolojilerinin edebi analiz alanına entegrasyonu, yeni boyutları ortaya çıkarmayı, yazılı içerik anlayışımızı zenginleştirmeyi ve edebiyat araştırmalarında yenilikçi uygulamaların önünü açmayı vaat ediyor. Bu makale aracılığıyla kitap özeti sınıflandırması ve tahmini alanında kelime gömme teknikleri ve makine öğreniminin sinerjik potansiyeline ilişkin kapsamlı bir araştırma sunulmuştur.

## 2 Yöntemler

Bu çalışmada yöntemler ve kullanılan metotlar alt başlıklar halinde ele alınmıştır. Bu aşamalar veri setinin toplanması ve hazırlanması, önişleme işlemleri, kelime gömme tekniklerinin uygulanması, makine öğrenmesi modellerinin kullanılması, model ve kelime gömme tekniklerinin karşılaştırılmasından oluşur.

### *2.1 Veri setinin hazırlanması ve Önişleme*

Kitap özetleri ve bilgilerini içeren veri seti, üç bin iki yüz bin veri ve sekiz kategoriden oluşmaktadır. Bu kategoriler "fantastik, bilim kurgu, romantik, tarih, polisiye, felsefe, sinema ve korku-gerilim" alt başlıklarından oluşmaktadır. Veriler popüler kitap satın alma sitesi olan idefix.com web sitesinden alınmıştır. Veri setinde sınıflandırma için kullanılan öznitelikler kitap özetleri ve kitap türleri sütunlarıdır.

Verilerin toplanmasının ardından verilerin daha iyi işlenmesi için doğal dil işlemede kullanılan bazı temel ön işleme metotları kullanılmıştır. Bu metotlar küçük harf dönüşümü, noktalama işaretlerinin, sayıların ve alfanümerik karakterlerin kaldırılması, kelimelerin köklerine ayırma ve duraksama kelimelerinin kaldırılmasından oluşmaktadır.

Küçük harf dönüşümü içeren veri setlerinde tüm kelimeler küçük harflere dönüştürülmüş ve makinenin büyük harfle başlayan kelimelerle karıştırmasının önüne geçilmiştir.

Noktalama işaretlerinin kaldırılmasını içeren veri setlerinde tüm noktalama işaretleri, alfanümerik karakterler ve sayılar sınıflandırma probleminde anlamlı olmadığından dolayı kaldırılmıştır.

Köklere ayırma yöntemi içeren veri setlerinde modelin başarısını arttırmak için kelimeler köklere indirgenmiştir. Köklere ayırma yöntemi için Turkish Stemmer for Python projesi Tuncelli ve Özdemir (2019) tarafından geliştirilmiştir ve Turkish Stemmer kök ayırma yöntemi olarak kullanılmıştır [7]. Ek olarak duraksama kelimelerinin (stopwords) makine öğreniminde anlamı olmadığı için Türkçedeki duraksama kelimeleri veri setinden kaldırılmıştır.

Ön işleme metotları aynı veri seti üzerinde farklı kombinasyonlarla uygulanıp kelime gömme teknikleri ve sınıflandırma yapılması için ayrılmıştır. Ön işleme tekniklerinin daha iyi gösterilmesi ve veri setinde uygulanan kombinasyonlarının karşılaştırılması için Tablo 1'de görülen şekilde kodlamalar oluşturulmuş ve sonuçlar kısmında kullanılmıştır.

Tablo 1. Önişleme Yöntemlerinin Kombinasyonları.

| Senaryo Numarası | Kod | Küçük Harf Dönüşümü Yok (0)/Var (1) | Noktalama İşaretlerinin Kaldırılması Yok (0)/ Var (1) | Köklere Ayırma Yok (0)/ Var (1) | Duraksama Kelimelerinin Kaldırılması Yok (0) /Var (1) |
|---|---|---|---|---|---|
| 1 | 0000 | 0 | 0 | 0 | 0 |
| 2 | 0001 | 0 | 0 | 0 | 1 |
| 3 | 0010 | 0 | 0 | 1 | 0 |
| 4 | 0011 | 0 | 0 | 1 | 1 |
| 5 | 0100 | 0 | 1 | 0 | 0 |
| 6 | 0101 | 0 | 1 | 0 | 1 |
| 7 | 0110 | 0 | 1 | 1 | 0 |
| 8 | 0111 | 0 | 1 | 1 | 1 |
| 9 | 1000 | 1 | 0 | 0 | 0 |
| 10 | 1001 | 1 | 0 | 0 | 1 |
| 11 | 1010 | 1 | 0 | 1 | 0 |
| 12 | 1011 | 1 | 0 | 1 | 1 |
| 13 | 1100 | 1 | 1 | 0 | 0 |
| 14 | 1101 | 1 | 1 | 0 | 1 |
| 15 | 1110 | 1 | 1 | 1 | 0 |
| 16 | 1111 | 1 | 1 | 1 | 1 |

### 2.2 Kelime Gömme Teknikleri

Günümüzde, doğal dil işleme ve yapay zeka alanlarındaki ilerlemeler kelime gömme tekniklerine olan ilgiyi artırmıştır. Kelime gömme, bir kelimenin bir vektör uzayında temsil edilmesidir. Bu teknikler bir kelimenin anlamını daha iyi anlamak, benzer anlamlı kelimeleri bulmak, dil modellerini eğitmek ve doğal dil işleme çalışmalarında başarıyı arttırmak için kullanılır. Bu makalede, kelime gömme tekniklerinden One Hot Encoding, Word2Vec ve Terim-Frekansı – Ters Doküman Frekansı (TF-IDF) yöntemleri sınıflandırma başarısını arttırmak ve model performanslarını ölçmek için kullanılmıştır.

### 2.2.1 One Hot Encoding

One hot encoding, kategorik verileri işlemek için yaygın olarak kullanılan bir kelime gömme tekniğidir. Bu teknik, kategorik değişkenleri her bir kategori için ayrı birer sütun olacak şekilde ikili vektörlerle temsil etmesini sağlar. Bir veri noktası bir kategoriye aitse o kategori için ilgili sütun 1'e, diğer tüm sütunlar ise 0'a ayarlanır.

One hot encoding, özellikle makine öğrenmesi algoritmalarında kategorik verilerle çalışırken sıklıkla kullanılır. Ancak çok sayıda farklı kategori varsa ve her bir kategori için bir sütun oluşturmak gerekiyorsa, bu teknik veri setinin boyutunu büyük ölçüde artırabilir ve bu da veri setinin işlenmesini zorlaştırabilir.

### 2.2.2 Terim Frekansı - Ters Doküman Frekans

Metin verilerini daha iyi anlamak ve içerdikleri bilgileri çıkarmak için çeşitli istatistiksel yöntemler kullanılır. Bu yöntemlerden biri de Terim Frekansı – Ters Doküman Frekansı (TF-IDF) adı verilen kelime gömme yöntemidir. Bu yöntemdeki kısımlardan Terim Frekansı (TF) bir belgedeki bir kelimenin frekansını belirten bir ölçüdür. Genellikle bir kelimenin bir belgedeki toplam kelime sayısına bölünerek hesaplanır. Bu, bir kelimenin belgedeki ne kadar sık geçtiğini gösterir. Ters Doküman Frekansı (IDF) ise bir kelimenin belirli bir belge koleksiyonundaki nadirlik derecesini belirten bir ölçüdür. IDF, belge koleksiyonundaki belge sayısının, belirli bir kelimenin geçtiği belge sayısının logaritması alınarak hesaplanır. Bu, bir kelimenin ne kadar nadir veya yaygın olduğunu gösterir.

TF-IDF, metin verilerini analiz etmek için çeşitli uygulamalarda kullanılır. Özellikle belge sınıflandırma ve bilgi çıkarma gibi alanlarda önemli bir rol oynar. Yüksek TF-IDF skorlarına sahip kelimeler, belgenin içeriğindeki önemli kavramları ve anahtar bilgileri temsil eder.

$$W_{a,b} = tf_{a,b} \times \log \frac{N}{df_a} \quad (1)$$

$tf_{a,b}$ = a kelimesinin b belgesindeki frekansı

$df_a$ = a kelimesini içeren döküman sayısı

N = Toplam doküman sayısı

### 2.1.3 Word2Vec

Word2Vec, doğal dil işleme alanında kelime gömme yöntemlerinden biridir ve kelime temsillerini vektör uzayında oluşturur. Bu yöntem, bir kelimenin anlamını ve ilişkilerini temsil etmek için kullanılır. Word2Vec modeli, büyük metin veri setlerinden öğrenilir ve ardından her kelimeyi yoğun bir vektörle temsil eder.

Word2Vec'in temel amacı, bir kelimenin anlamını çevresindeki kelimelerle ilişkilendirerek vektör uzayında temsil etmektir. Bu ilişki, bir kelimenin belirli bir bağlam içindeki olasılığına dayanır. Bu yöntem ilk olarak Mikolov ve arkadaşları tarafından önerilmiştir [5]. Word2Vec'in iki alt yöntemi vardır. Bunlar Continous Bag of Words (CBOW) ve Skip-Gram yöntemleridir.

CBOW yöntemi bir kelimenin çevresindeki diğer kelimeler kullanılarak o kelimenin tahmin edilmesine dayanır. Yani, bir kelimeye bakılarak etrafındaki kelimelerin olasılıkları hesaplanır. Skip-gram yöntemi ise tam tersine, bir kelimenin verildiğinde bu kelimenin çevresindeki diğer kelimelerin tahmin edilmesine odaklanır. Yani, bir kelime verildiğinde, etrafındaki kelimelerin olasılıklarını tahmin eder. Şekil 1'de CBOW ve Skip-Gram yöntemlerinin çalışma yapıları gösterilmiştir. Word2Vec ek olarak büyük verilerin işlenmesi için de kullanılabilir. Yapılan bir çalışmada Word2Vec yöntemlerinden olan CBOW ve Skip-Gram yöntemleri büyük veriler üzerinde kullanılmış ve büyük verilerin boyutlarının azaltılmasında bir strateji olarak uygulanmıştır [6].

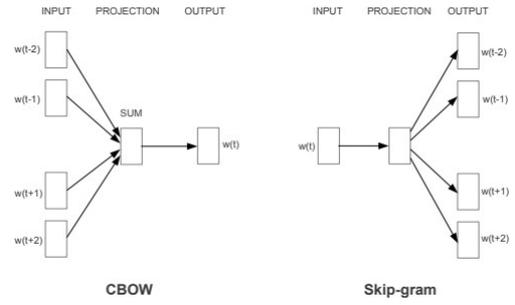

Şekil 1. CBOW ve Skip-Gram Yapısı

## 3 Makine Öğrenmesi Modelleri

Ön işleme ve kelime gömme tekniklerinin kullanılmasının ardından temel makine öğrenmesi modelleri kitap özetlerinin 8 kategoride sınıflandırılması için kullanılmıştır. Bu modeller K-En Yakın Komşu Algoritması (KNN), Naive Bayes (NB), Rastgele Karar Ormanı (RF), Karar Ağaçları (DT), Destek Vektör Makineleri (SVM), Lojistik Regresyon (LR) ve AdaBoost (AB) modelleridir.

### 3.1 K-En Yakın Komşu (KNN)

K-En Yakın Komşu (KNN), makine öğrenmesinde sınıflandırma ve regresyon problemlerini çözmek için kullanılan basit ancak güçlü bir algoritmadır. 1967 yılında T. M. Cover ve P. E. Hart tarafından önerilmiştir [8]. Bu algoritma, örneklerin birbirlerine olan benzerliklerine dayanarak tahminler yapar. KNN algoritması, basit bir prensibe dayanır: "Bir veri noktasını sınıflandırmak veya tahmin etmek için, o noktaya en yakın olan k komşulardan çoğunluğunu al." Burada "k" bir hiperparametredir ve kullanıcı tarafından belirlenir. Bu hiperparametre, komşuluk boyutunu belirler ve tahminlerin doğruluğunu etkiler.

Bir veri noktasının sınıflandırılması gerektiğinde, bu noktaya en yakın k komşu bulunur. Ardından, bu komşuların etiketleri kullanılarak sınıflandırma yapılır veya komşuların değerleri ortalama alınarak regresyon yapılır.

### 3.2 Naive Bayes (NB)

Naive Bayes (NB), makine öğrenmesinde sınıflandırma problemlerini çözmek için kullanılan olasılık temelli bir algoritmadır. Bu algoritma, Bayes teoremi prensibine dayanır ve özellikler arasındaki bağımsızlık varsayımını kabul eder. Sınıflandırma problemlerinden ikili sınıflandırmalarda, çoklu kategorik sınıflandırmalarda hatta dengelenmemiş sınıflar için de Naive Bayes makine öğrenmesi modeli kullanılmaktadır [9]. Naive Bayes algoritması, bir veri noktasının sınıfını belirlemek için özellikler arasındaki olasılıkları kullanır. Bu algoritma, Bayes teoremi prensibine dayanır ve özellikler arasındaki bağımsızlık varsayımını kabul eder. Bu varsayıma göre, bir veri noktasının sınıfı, bu noktanın özelliklerinin sınıflar arasındaki olasılıkları kullanılarak belirlenir.

$$P\left(\frac{A}{B}\right) = \frac{P\left(\frac{B}{A}\right) \cdot P(A)}{P(B)} \quad (2)$$

P(A|B) = B olayının gerçekleştiği durumda A olayının gerçekleşme olasılığını ifade eder.

P(A) = A olayının gerçekleşme olasılığını belirtir.

P(B|A) = A olayının gerçekleştiği durumda B olayının gerçekleşme olasılığını ifade eder.

P(B) = B olayının gerçekleşme olasılığını ifade eder.

### 3.3 Rastgele Karar Ormanı (RF)

Rastgele Karar Ormanı algoritması, makine öğrenmesinde sınıflandırma ve regresyon problemlerini çözmek için kullanılan bir öğrenme algoritmasıdır. Bu algoritma, birden fazla karar ağacının oluşturulması ve bunların bir araya getirilmesiyle çalışır. Her karar ağacı, rastgele örneklem alınan veri noktalarıyla ve rastgele özelliklerle eğitilir. Ardından, her ağaç bir sınıf tahmini yapar ve bu tahminlerin çoğunluğu alınarak modelin tahmini belirlenir. Bu algoritma ayrıca geniş bir uygulama alanına sahiptir ve birçok endüstriyel ve akademik problemin çözümünde başarıyla kullanılmaktadır. Özellikle, büyük veri kümeleri ve yüksek boyutlu özellik uzaylarıyla çalışırken iyi performans gösterir. İngilizce haber metinlerinin sınıflandırılmasında klasik makine öğrenmesi modelleri ile karşılaştırıldığında Rastgele Karar Ormanı algoritması %0.93'lük bir başarı oranı ile öne çıkmıştır [10].

### 3.4 Karar Ağaçları (DT)

Karar ağaçları, bir veri kümesindeki özelliklerin değerlerine göre sınıfları veya değerleri tahmin etmek için kullanılan bir modelleme tekniğidir. Bir karar ağacı, veri kümesini bir dizi karar düğümünde bölerek sınıflara veya değerlere yönlendirir. Her bir karar düğümü, bir özelliğin değerini kontrol eder ve bu değere göre veriyi iki veya daha fazla alt kümeye ayırır. Bu bölünmeler, veri kümesinin daha küçük ve daha homojen alt gruplara ayrılmasını sağlar.

### 3.5 Destek Vektör Makinesi (SVM)

Destek vektör makineleri, sınıflandırma ve regresyon problemleri için kullanılan bir makine öğrenmesi algoritmasıdır. SVM, veri noktalarını sınıflandırmak için bir hiperdüzlem oluşturur. Bu hiperdüzlem, veri noktalarını iki veya daha fazla sınıfa ayırmak için en iyi şekilde ayıran bir çizgidir. SVM, sınıflar arasındaki marjı maksimize etmek için çalışır, yani en yakın veri noktaları arasındaki mesafeyi maksimize eder. SVM, lineer ve doğrusal olmayan sınıflandırma problemlerini çözebilir ve yüksek boyutlu veri kümelerinde etkilidir. Spam sms mesajlarının sınıflandırılması üzerine yapılan bir çalışmada klasik makine öğrenmesi modelleri karşılaştırılmış ve sonuçlara bakıldığında SVM modeli ortalama 98.9% oranında bir başarıyla spam mesajlarını tespit etmede öne geçmiştir [11]. Ayrıca SVM, çeşitli çekirdek fonksiyonları kullanarak doğrusal olmayan sınıflandırma problemlerinin çözümlerinde kullanılabilir. Yapılan başka bir çalışmada farklı değişken sayıları ile metin sınıflandırılması yapılmış ve klasik makine öğrenmesi modelleri ile karşılaştırılarak SVM modeli daha başarılı olmuştur [12].

### 3.6 Lojistik Regresyon (LR)

Lojistik regresyon, bir bağımlı değişkenin olasılık dağılımını, bağımsız değişkenlerin bir lineer kombinasyonu ile tahmin etmek için kullanılan bir istatistiksel modeldir. Lojistik regresyon, sınıflandırma problemleri için yaygın olarak kullanılır. İkili sınıflandırma problemlerinde, lojistik regresyon çıktıyı 0 ile 1 arasında bir olasılık değeri olarak verir. Bu olasılık değeri, bir veri noktasının bir sınıfa ait olma olasılığını belirtir. Lojistik regresyon, lineer olmayan ilişkileri modellemek için kullanılan lineer bir modeldir. Lojistik regresyon ayrıca birçok kelime gömme tekniği ile uyumlu olarak kullanılabilir. Sosyal medya yorumları ile yapılan duygu analizi çalışmasında lojistik regresyon modeli, kelime gömme tekniklerinden kelime çantası tekniği (bag of words) için en başarılı sonucu verirken TF-IDF tekniği için ise ikinci en başarılı sonucu vermiştir [13]. Ayrıca duygu sınıflandırılması için yapılan başka bir çalışmada derin öğrenme modellerinden Uzun-Kısa Süreli Bellek (LSTM) modeli ve lojistik regresyon karşılaştırılmış ve lojistik regresyon modeli hem Twitter hem de IMDB veri setleri için daha başarılı sonuçlar vermiştir [14].

### 3.7 AdaBoost (AB)

AdaBoost, genellikle karar ağaçlarını bir araya getirerek güçlü bir sınıflandırıcı oluşturan bir öğrenme algoritmasıdır. AdaBoost, her bir öğreniciyi

16

sırayla eğitir ve daha fazla vurgu verilen yanlış sınıflandırılmış örnekleri belirler. Daha sonra, yanlış sınıflandırılmış örneklerin üzerine daha fazla vurgu yaparak yeni bir öğrenici eğitir. Bu şekilde, AdaBoost, daha fazla vurgu gerektiren örnekler üzerinde daha güçlü bir öğrenme gerçekleştirir. AdaBoost, yüksek doğruluk sağlayabilir ve genellikle diğer sınıflandırma algoritmalarıyla birlikte kullanıldığında performansını artırabilir. Dengesiz bir veri seti için yapılan sınıflandırma çalışmasında AdaBoost algoritması performansı arttırmak için önerilen bir metotla kullanılmış ve başarılı sonuç vermiştir [15].

## *4 Bulgular ve Yorumlar*

Bu çalışmada kitap özetleri veri seti ön işlemlerden geçirildikten sonra her bir kombinasyonu içeren veri setine farklı kelime gömme teknikleri uygulanmıştır. Daha sonra kelime gömme teknikleri uygulanan her bir veri seti kombinasyonu için temel makine öğrenmesi modelleri sınıflandırma yapılmak üzere uygulanmış ve Tablo 3'te detaylı bir şekilde gösterilmiştir. Ayrıca en başarılı makine öğrenmesi modeline göre oluşturulan karşılaştırmalar kelime gömme teknikleri ile birlikte Tablo 2'de gösterilmiştir. Değerlendirme kriterleri olarak çalışmada 4 adet parametre kullanılmıştır: F-Skor (FS), Doğruluk (ACC), Precision (PRC) ve Recall (RC). Bulgulara bakıldığında ön işleme metotlarından köklerine ayırma yönteminin bütün kelime gömme teknikleri için model başarısını arttırdığı görülmektedir. TF-IDF tekniği uygulanan modeller için en yüksek başarı ve F-Skor, SVM modelinde 0.8 olarak görülmüştür. Ayrıca diğer ön işleme kombinasyonlarında da SVM modeli diğer makine öğrenmesi modellerine göre daha iyi sonuç vermiştir. Word2Vec uygulanan veri setlerindeki başarı oranlarına bakıldığında lojistik regresyon modeli ve SVM diğer öğrenme modellerinden daha başarılı sonuçlar verdiği gözlenmiştir. En yüksek başarı ve F-Skor 0010 kombinasyonu için 0.72 olarak LR modelinde görülmüştür. Bunun nedenleri arasında lojistik regresyon modelinin girdi olarak Wod2Vec ile oluşturulan kelime vektörlerini alarak metinlerin daha iyi bir şekilde temsil edilmesini sağlaması gösterilebilir. Bu, modelin daha iyi özellik temsili elde etmesine ve dolayısıyla daha iyi performans göstermesine olanak tanır. Son olarak One-Hot Encoding yöntemi kullanılan veri setleri üzerinde en başarılı sonuç ve F-Skor 1010 ve 1011 kombinasyonları için Naive Bayes modelinde 0.81 olarak görülmüştür.

Tablo 2. Kelime Gömme Teknikleri ve Ön İşleme Kombinasyonlarının En Başarılı Modeller ile Karşılaştırılması

|      | TF-IDF | | | Word2Vec | | | One Hot Encoding | | |
|------|------|------|-------|------|------|-------|------|------|-------|
| Kod  | FS   | ACC  | Model | FS   | ACC  | Model | FS   | ACC  | Model |
| 0000 | 0.78 | 0.77 | SVM   | 0.68 | 0.68 | LR    | 0.79 | 0.78 | NB    |
| 0001 | 0.79 | 0.79 | SVM   | 0.67 | 0.67 | LR    | 0.8  | 0.8  | NB    |
| 0010 | 0.79 | 0.79 | SVM   | 0.71 | 0.72 | LR    | 0.8  | 0.8  | LR    |
| 0011 | 0.8  | 0.8  | SVM   | 0.7  | 0.71 | LR    | 0.8  | 0.8  | LR    |
| 0100 | 0.77 | 0.77 | SVM   | 0.68 | 0.68 | LR    | 0.78 | 0.78 | NB    |
| 0101 | 0.77 | 0.77 | SVM   | 0.68 | 0.68 | LR    | 0.78 | 0.78 | NB    |
| 0110 | 0.79 | 0.79 | SVM   | 0.7  | 0.7  | LR    | 0.79 | 0.78 | LR    |
| 0111 | 0.79 | 0.79 | SVM   | 0.7  | 0.7  | LR    | 0.79 | 0.78 | LR    |
| 1000 | 0.78 | 0.77 | SVM   | 0.67 | 0.68 | LR    | 0.79 | 0.78 | NB    |
| 1001 | 0.79 | 0.78 | SVM   | 0.67 | 0.67 | SVM   | 0.79 | 0.79 | NB    |
| 1010 | 0.8  | 0.8  | SVM   | 0.69 | 0.7  | LR    | 0.81 | 0.81 | NB    |
| 1011 | 0.8  | 0.8  | SVM   | 0.7  | 0.7  | LR    | 0.81 | 0.81 | NB    |
| 1100 | 0.77 | 0.77 | SVM   | 0.67 | 0.68 | LR    | 0.78 | 0.78 | NB    |
| 1101 | 0.79 | 0.78 | SVM   | 0.68 | 0.68 | LR    | 0.78 | 0.78 | NB-LR |
| 1110 | 0.79 | 0.79 | SVM   | 0.69 | 0.69 | LR    | 0.79 | 0.8  | LR    |
| 1111 | 0.79 | 0.79 | SVM   | 0.7  | 0.7  | LR    | 0.8  | 0.8  | NB-LR |

Tablo 3. One-Hot Encoding, TF-IDF, Word2Vec Modelleri ve Ön İşleme Kombinasyonlarının Makine Öğrenmesi Modelleri ile Karşılaştırılması

| | Algoritmalar | Değerlendirme Kriterleri |
|---|---|---|

17

| Önişlem Kodu | | One-Hot Encoding | | | | TF-IDF | | | | Word2Vec | | | |
|---|---|---|---|---|---|---|---|---|---|---|---|---|---|
| | | FS | ACC | PRC | RC | FS | ACC | PRC | RC | FS | ACC | PRC | RC |
| 0000 | KNN [k=15] | 0.16 | 0.07 | 0.33 | 0.16 | 0.069 | 0.14 | 0.32 | 0.14 | 0.54 | 0.54 | 0.57 | 0.54 |
| | NB | 0.79 | 0.78 | 0.8 | 0.78 | 0.75 | 0.75 | 0.77 | 0.76 | 0.48 | 0.48 | 0.51 | 0.47 |
| | RF | 0.74 | 0.74 | 0.75 | 0.74 | 0.69 | 0.68 | 0.71 | 0.69 | 0.57 | 0.57 | 0.57 | 0.57 |
| | DT | 0.59 | 0.59 | 0.6 | 0.59 | 0.55 | 0.54 | 0.55 | 0.55 | 0.38 | 0.38 | 0.39 | 0.38 |
| | SVM | 0.71 | 0.72 | 0.72 | 0.72 | 0.78 | 0.77 | 0.78 | 0.78 | 0.67 | 0.67 | 0.68 | 0.67 |
| | LR | 0.78 | 0.78 | 0.78 | 0.78 | 0.76 | 0.75 | 0.77 | 0.76 | 0.68 | 0.68 | 0.68 | 0.68 |
| | AB | 0.43 | 0.43 | 0.79 | 0.38 | 0.46 | 0.46 | 0.73 | 0.43 | 0.31 | 0.31 | 0.35 | 0.33 |
| 0001 | KNN [k=15] | 0.16 | 0.07 | 0.33 | 0.16 | 0.08 | 0.14 | 0.32 | 0.15 | 0.55 | 0.55 | 0.58 | 0.55 |
| | NB | 0.8 | 0.8 | 0.82 | 0.79 | 0.77 | 0.77 | 0.78 | 0.77 | 0.49 | 0.49 | 0.52 | 0.48 |
| | RF | 0.75 | 0.75 | 0.76 | 0.75 | 0.69 | 0.68 | 0.71 | 0.69 | 0.6 | 0.6 | 0.61 | 0.6 |
| | DT | 0.59 | 0.59 | 0.59 | 0.59 | 0.55 | 0.54 | 0.56 | 0.55 | 0.36 | 0.36 | 0.37 | 0.36 |
| | SVM | 0.71 | 0.72 | 0.72 | 0.72 | 0.79 | 0.79 | 0.79 | 0.79 | 0.66 | 0.66 | 0.67 | 0.66 |
| | LR | 0.78 | 0.78 | 0.78 | 0.78 | 0.77 | 0.76 | 0.78 | 0.77 | 0.67 | 0.67 | 0.68 | 0.67 |
| | AB | 0.43 | 0.43 | 0.79 | 0.38 | 0.45 | 0.45 | 0.74 | 0.42 | 0.31 | 0.31 | 0.32 | 0.36 |
| 0010 | KNN [k=15] | 0.16 | 0.07 | 0.33 | 0.16 | 0.07 | 0.14 | 0.31 | 0.14 | 0.6 | 0.6 | 0.62 | 0.6 |
| | NB | 0.79 | 0.79 | 0.81 | 0.79 | 0.76 | 0.76 | 0.79 | 0.77 | 0.52 | 0.52 | 0.55 | 0.51 |
| | RF | 0.74 | 0.75 | 0.76 | 0.75 | 0.71 | 0.71 | 0.72 | 0.72 | 0.59 | 0.59 | 0.6 | 0.59 |
| | DT | 0.61 | 0.61 | 0.61 | 0.61 | 0.55 | 0.55 | 0.55 | 0.55 | 0.38 | 0.38 | 0.38 | 0.37 |
| | SVM | 0.75 | 0.75 | 0.75 | 0.75 | 0.79 | 0.79 | 0.8 | 0.8 | 0.68 | 0.68 | 0.68 | 0.67 |
| | LR | 0.8 | 0.8 | 0.8 | 0.8 | 0.77 | 0.77 | 0.79 | 0.78 | 0.71 | 0.72 | 0.72 | 0.71 |
| | AB | 0.48 | 0.48 | 0.67 | 0.45 | 0.41 | 0.41 | 0.55 | 0.43 | 0.28 | 0.28 | 0.3 | 0.31 |
| 0011 | KNN [k=15] | 0.06 | 0.15 | 0.22 | 0.14 | 0.07 | 0.14 | 0.31 | 0.15 | 0.6 | 0.6 | 0.62 | 0.59 |
| | NB | 0.79 | 0.79 | 0.81 | 0.79 | 0.78 | 0.77 | 0.79 | 0.78 | 0.53 | 0.53 | 0.56 | 0.52 |
| | RF | 0.75 | 0.76 | 0.76 | 0.76 | 0.73 | 0.73 | 0.75 | 0.74 | 0.61 | 0.61 | 0.61 | 0.6 |
| | DT | 0.6 | 0.6 | 0.6 | 0.6 | 0.56 | 0.56 | 0.57 | 0.57 | 0.39 | 0.39 | 0.41 | 0.39 |
| | SVM | 0.76 | 0.76 | 0.76 | 0.76 | 0.8 | 0.8 | 0.8 | 0.8 | 0.68 | 0.68 | 0.68 | 0.67 |
| | LR | 0.8 | 0.8 | 0.8 | 0.8 | 0.79 | 0.78 | 0.8 | 0.79 | 0.7 | 0.71 | 0.71 | 0.7 |
| | AB | 0.48 | 0.48 | 0.67 | 0.45 | 0.5 | 0.5 | 0.59 | 0.5 | 0.29 | 0.29 | 0.31 | 0.31 |
| 0100 | KNN [k=15] | 0.08 | 0.17 | 0.2 | 0.17 | 0.07 | 0.14 | 0.31 | 0.15 | 0.52 | 0.53 | 0.55 | 0.52 |
| | NB | 0.78 | 0.78 | 0.8 | 0.78 | 0.74 | 0.74 | 0.77 | 0.75 | 0.46 | 0.46 | 0.5 | 0.46 |
| | RF | 0.74 | 0.75 | 0.76 | 0.75 | 0.68 | 0.67 | 0.7 | 0.68 | 0.56 | 0.56 | 0.57 | 0.56 |
| | DT | 0.58 | 0.58 | 0.58 | 0.58 | 0.55 | 0.54 | 0.55 | 0.54 | 0.34 | 0.34 | 0.34 | 0.34 |
| | SVM | 0.71 | 0.71 | 0.71 | 0.71 | 0.77 | 0.77 | 0.78 | 0.78 | 0.66 | 0.66 | 0.67 | 0.65 |
| | LR | 0.77 | 0.77 | 0.77 | 0.77 | 0.75 | 0.75 | 0.77 | 0.76 | 0.68 | 0.68 | 0.68 | 0.68 |
| | AB | 0.44 | 0.44 | 0.77 | 0.4 | 0.34 | 0.34 | 0.69 | 0.34 | 0.3 | 0.3 | 0.37 | 0.31 |
| 0101 | KNN [k=15] | 0.08 | 0.17 | 0.2 | 0.17 | 0.07 | 0.14 | 0.31 | 0.15 | 0.53 | 0.53 | 0.56 | 0.53 |
| | NB | 0.78 | 0.78 | 0.8 | 0.78 | 0.74 | 0.74 | 0.77 | 0.75 | 0.46 | 0.46 | 0.5 | 0.46 |
| | RF | 0.73 | 0.73 | 0.75 | 0.73 | 0.67 | 0.67 | 0.69 | 0.68 | 0.55 | 0.55 | 0.56 | 0.56 |
| | DT | 0.59 | 0.59 | 0.6 | 0.59 | 0.55 | 0.54 | 0.55 | 0.55 | 0.35 | 0.35 | 0.36 | 0.35 |
| | SVM | 0.71 | 0.71 | 0.71 | 0.71 | 0.77 | 0.77 | 0.78 | 0.78 | 0.66 | 0.66 | 0.67 | 0.65 |
| | LR | 0.77 | 0.77 | 0.77 | 0.77 | 0.75 | 0.75 | 0.77 | 0.76 | 0.68 | 0.68 | 0.68 | 0.68 |
| | AB | 0.44 | 0.44 | 0.77 | 0.4 | 0.34 | 0.34 | 0.69 | 0.34 | 0.3 | 0.3 | 0.37 | 0.31 |
| 0110 | KNN [k=15] | 0.08 | 0.18 | 0.26 | 0.17 | 0.06 | 0.14 | 0.31 | 0.14 | 0.58 | 0.58 | 0.6 | 0.58 |
| | NB | 0.79 | 0.78 | 0.8 | 0.79 | 0.75 | 0.74 | 0.78 | 0.75 | 0.52 | 0.52 | 0.54 | 0.51 |
| | RF | 0.73 | 0.74 | 0.75 | 0.74 | 0.7 | 0.7 | 0.72 | 0.71 | 0.59 | 0.59 | 0.59 | 0.58 |
| | DT | 0.6 | 0.6 | 0.6 | 0.6 | 0.54 | 0.53 | 0.54 | 0.54 | 0.38 | 0.38 | 0.38 | 0.37 |
| | SVM | 0.75 | 0.75 | 0.75 | 0.75 | 0.79 | 0.79 | 0.8 | 0.79 | 0.67 | 0.67 | 0.68 | 0.66 |
| | LR | 0.8 | 0.8 | 0.8 | 0.8 | 0.77 | 0.77 | 0.79 | 0.78 | 0.7 | 0.7 | 0.71 | 0.7 |
| | AB | 0.42 | 0.42 | 0.53 | 0.43 | 0.49 | 0.49 | 0.57 | 0.5 | 0.31 | 0.31 | 0.35 | 0.33 |
| 0111 | KNN [k=15] | 0.08 | 0.18 | 0.26 | 0.17 | 0.06 | 0.14 | 0.31 | 0.14 | 0.58 | 0.58 | 0.6 | 0.58 |
| | NB | 0.79 | 0.78 | 0.8 | 0.79 | 0.75 | 0.74 | 0.78 | 0.75 | 0.52 | 0.52 | 0.54 | 0.51 |
| | RF | 0.74 | 0.74 | 0.75 | 0.74 | 0.69 | 0.69 | 0.72 | 0.7 | 0.59 | 0.59 | 0.59 | 0.59 |
| | DT | 0.6 | 0.6 | 0.6 | 0.6 | 0.55 | 0.54 | 0.55 | 0.55 | 0.38 | 0.38 | 0.39 | 0.37 |
| | SVM | 0.75 | 0.75 | 0.75 | 0.75 | 0.79 | 0.79 | 0.8 | 0.79 | 0.67 | 0.67 | 0.68 | 0.66 |
| | LR | 0.8 | 0.8 | 0.8 | 0.8 | 0.77 | 0.77 | 0.79 | 0.78 | 0.7 | 0.7 | 0.71 | 0.7 |
| | AB | 0.42 | 0.42 | 0.53 | 0.43 | 0.49 | 0.49 | 0.57 | 0.5 | 0.31 | 0.31 | 0.35 | 0.33 |
| 1000 | KNN [k=15] | 0.07 | 0.16 | 0.33 | 0.16 | 0.06 | 0.14 | 0.31 | 0.14 | 0.54 | 0.54 | 0.57 | 0.54 |
| | NB | 0.79 | 0.78 | 0.8 | 0.78 | 0.75 | 0.75 | 0.77 | 0.76 | 0.48 | 0.48 | 0.51 | 0.47 |
| | RF | 0.74 | 0.74 | 0.76 | 0.75 | 0.7 | 0.69 | 0.72 | 0.7 | 0.57 | 0.57 | 0.57 | 0.56 |
| | DT | 0.59 | 0.59 | 0.59 | 0.59 | 0.54 | 0.54 | 0.54 | 0.54 | 0.34 | 0.35 | 0.34 | 0.34 |
| | SVM | 0.71 | 0.72 | 0.72 | 0.72 | 0.78 | 0.77 | 0.78 | 0.78 | 0.67 | 0.67 | 0.68 | 0.67 |
| | LR | 0.78 | 0.78 | 0.78 | 0.78 | 0.76 | 0.75 | 0.77 | 0.76 | 0.67 | 0.68 | 0.68 | 0.67 |
| | AB | 0.43 | 0.43 | 0.79 | 0.38 | 0.46 | 0.46 | 0.73 | 0.43 | 0.31 | 0.31 | 0.35 | 0.33 |
| 1001 | KNN [k=15] | 0.07 | 0.16 | 0.13 | 0.16 | 0.07 | 0.14 | 0.31 | 0.15 | 0.54 | 0.54 | 0.57 | 0.54 |
| | NB | 0.79 | 0.79 | 0.8 | 0.79 | 0.77 | 0.76 | 0.78 | 0.77 | 0.48 | 0.49 | 0.52 | 0.48 |
| | RF | 0.76 | 0.76 | 0.77 | 0.76 | 0.69 | 0.68 | 0.71 | 0.69 | 0.59 | 0.59 | 0.6 | 0.59 |
| | DT | 0.57 | 0.58 | 0.58 | 0.58 | 0.55 | 0.54 | 0.55 | 0.55 | 0.35 | 0.35 | 0.36 | 0.35 |
| | SVM | 0.71 | 0.72 | 0.72 | 0.71 | 0.79 | 0.78 | 0.79 | 0.79 | 0.67 | 0.67 | 0.68 | 0.67 |
| | LR | 0.78 | 0.79 | 0.79 | 0.79 | 0.77 | 0.76 | 0.78 | 0.77 | 0.67 | 0.67 | 0.68 | 0.67 |
| | AB | 0.43 | 0.43 | 0.79 | 0.38 | 0.44 | 0.44 | 0.71 | 0.41 | 0.33 | 0.33 | 0.34 | 0.36 |
| 1010 | KNN [k=15] | 0.12 | 0.22 | 0.32 | 0.22 | 0.07 | 0.14 | 0.31 | 0.15 | 0.57 | 0.57 | 0.6 | 0.57 |
| | NB | 0.81 | 0.81 | 0.83 | 0.8 | 0.76 | 0.76 | 0.78 | 0.77 | 0.51 | 0.51 | 0.54 | 0.51 |
| | RF | 0.75 | 0.76 | 0.76 | 0.75 | 0.7 | 0.7 | 0.74 | 0.71 | 0.59 | 0.59 | 0.6 | 0.59 |
| | DT | 0.61 | 0.61 | 0.61 | 0.61 | 0.56 | 0.56 | 0.56 | 0.57 | 0.38 | 0.38 | 0.39 | 0.38 |
| | SVM | 0.73 | 0.73 | 0.73 | 0.73 | 0.8 | 0.8 | 0.8 | 0.8 | 0.68 | 0.68 | 0.69 | 0.68 |
| | LR | 0.8 | 0.8 | 0.8 | 0.8 | 0.78 | 0.77 | 0.79 | 0.78 | 0.69 | 0.7 | 0.7 | 0.69 |
| | AB | 0.47 | 0.47 | 0.78 | 0.46 | 0.5 | 0.5 | 0.57 | 0.51 | 0.32 | 0.32 | 0.37 | 0.35 |
| 1011 | KNN [k=15] | 0.1 | 0.2 | 0.31 | 0.2 | 0.07 | 0.14 | 0.31 | 0.15 | 0.59 | 0.59 | 0.61 | 0.58 |
| | NB | 0.81 | 0.81 | 0.83 | 0.8 | 0.77 | 0.77 | 0.79 | 0.78 | 0.55 | 0.55 | 0.57 | 0.54 |
| | RF | 0.76 | 0.76 | 0.77 | 0.76 | 0.72 | 0.72 | 0.73 | 0.73 | 0.6 | 0.6 | 0.6 | 0.6 |
| | DT | 0.63 | 0.63 | 0.63 | 0.63 | 0.56 | 0.55 | 0.56 | 0.56 | 0.35 | 0.35 | 0.35 | 0.34 |
| | SVM | 0.74 | 0.74 | 0.75 | 0.74 | 0.8 | 0.8 | 0.81 | 0.81 | 0.67 | 0.67 | 0.68 | 0.67 |
| | LR | 0.8 | 0.8 | 0.8 | 0.8 | 0.79 | 0.79 | 0.8 | 0.79 | 0.7 | 0.7 | 0.71 | 0.7 |
| | AB | 0.47 | 0.47 | 0.78 | 0.46 | 0.49 | 0.49 | 0.56 | 0.49 | 0.3 | 0.3 | 0.32 | 0.32 |
| 1100 | KNN [k=15] | 0.08 | 0.17 | 0.2 | 0.17 | 0.07 | 0.14 | 0.31 | 0.15 | 0.53 | 0.53 | 0.56 | 0.53 |
| | NB | 0.78 | 0.78 | 0.8 | 0.78 | 0.74 | 0.74 | 0.77 | 0.75 | 0.46 | 0.46 | 0.5 | 0.46 |
| | RF | 0.73 | 0.74 | 0.75 | 0.74 | 0.66 | 0.66 | 0.68 | 0.67 | 0.57 | 0.57 | 0.57 | 0.57 |
| | DT | 0.58 | 0.58 | 0.58 | 0.58 | 0.54 | 0.53 | 0.54 | 0.53 | 0.35 | 0.35 | 0.36 | 0.35 |
| | SVM | 0.71 | 0.71 | 0.71 | 0.71 | 0.77 | 0.77 | 0.78 | 0.78 | 0.66 | 0.66 | 0.67 | 0.65 |
| | LR | 0.77 | 0.77 | 0.77 | 0.77 | 0.75 | 0.75 | 0.77 | 0.76 | 0.67 | 0.68 | 0.67 | 0.67 |
| | AB | 0.44 | 0.44 | 0.77 | 0.4 | 0.34 | 0.34 | 0.69 | 0.34 | 0.3 | 0.3 | 0.37 | 0.31 |
| 1101 | KNN [k=15] | 0.07 | 0.16 | 0.2 | 0.16 | 0.07 | 0.14 | 0.31 | 0.15 | 0.55 | 0.55 | 0.58 | 0.55 |
| | NB | 0.78 | 0.78 | 0.8 | 0.78 | 0.76 | 0.76 | 0.77 | 0.77 | 0.5 | 0.5 | 0.53 | 0.49 |
| | RF | 0.72 | 0.73 | 0.75 | 0.73 | 0.69 | 0.68 | 0.7 | 0.69 | 0.58 | 0.58 | 0.59 | 0.57 |
| | DT | 0.58 | 0.58 | 0.58 | 0.58 | 0.53 | 0.52 | 0.53 | 0.53 | 0.36 | 0.36 | 0.37 | 0.36 |
| | SVM | 0.72 | 0.72 | 0.72 | 0.72 | 0.79 | 0.78 | 0.79 | 0.79 | 0.67 | 0.67 | 0.68 | 0.67 |
| | LR | 0.78 | 0.78 | 0.78 | 0.78 | 0.76 | 0.76 | 0.77 | 0.77 | 0.68 | 0.68 | 0.69 | 0.68 |
| | AB | 0.44 | 0.44 | 0.77 | 0.4 | 0.39 | 0.39 | 0.76 | 0.38 | 0.33 | 0.33 | 0.33 | 0.36 |
| 1110 | KNN [k=15] | 0.11 | 0.21 | 0.26 | 0.2 | 0.07 | 0.14 | 0.31 | 0.15 | 0.58 | 0.58 | 0.61 | 0.58 |
| | NB | 0.79 | 0.79 | 0.8 | 0.79 | 0.76 | 0.76 | 0.79 | 0.77 | 0.52 | 0.52 | 0.55 | 0.51 |
| | RF | 0.76 | 0.77 | 0.78 | 0.77 | 0.7 | 0.69 | 0.73 | 0.7 | 0.59 | 0.59 | 0.59 | 0.58 |
| | DT | 0.6 | 0.6 | 0.6 | 0.6 | 0.59 | 0.58 | 0.59 | 0.59 | 0.38 | 0.38 | 0.39 | 0.37 |
| | SVM | 0.73 | 0.73 | 0.74 | 0.73 | 0.79 | 0.79 | 0.8 | 0.8 | 0.67 | 0.67 | 0.69 | 0.67 |
| | LR | 0.79 | 0.8 | 0.8 | 0.8 | 0.78 | 0.78 | 0.79 | 0.79 | 0.69 | 0.69 | 0.7 | 0.69 |
| | AB | 0.51 | 0.51 | 0.69 | 0.49 | 0.5 | 0.5 | 0.6 | 0.51 | 0.35 | 0.35 | 0.37 | 0.36 |
| 1111 | KNN [k=15] | 0.1 | 0.21 | 0.35 | 0.21 | 0.07 | 0.14 | 0.31 | 0.15 | 0.58 | 0.58 | 0.6 | 0.57 |
| | NB | 0.8 | 0.8 | 0.82 | 0.8 | 0.78 | 0.77 | 0.79 | 0.78 | 0.54 | 0.54 | 0.56 | 0.53 |
| | RF | 0.75 | 0.76 | 0.77 | 0.76 | 0.73 | 0.72 | 0.75 | 0.73 | 0.59 | 0.59 | 0.59 | 0.59 |
| | DT | 0.6 | 0.61 | 0.61 | 0.61 | 0.59 | 0.59 | 0.59 | 0.59 | 0.38 | 0.38 | 0.39 | 0.38 |
| | SVM | 0.74 | 0.74 | 0.74 | 0.74 | 0.79 | 0.79 | 0.8 | 0.8 | 0.67 | 0.67 | 0.68 | 0.67 |
| | LR | 0.8 | 0.8 | 0.8 | 0.8 | 0.78 | 0.78 | 0.79 | 0.79 | 0.7 | 0.7 | 0.71 | 0.7 |
| | AB | 0.51 | 0.51 | 0.69 | 0.49 | 0.5 | 0.5 | 0.59 | 0.51 | 0.32 | 0.32 | 0.34 | 0.34 |

## 5 Sonuçlar

Bu çalışmada farklı kelime gömme tekniklerinin, ön işleme kombinasyonlarının ve makine öğrenmesi modellerinin performansını karşılaştırdık. Tablo 2 ve Tablo 3'de görüldüğü üzere TF-IDF yöntemiyle kullanılan destek vektör makinesi modeli, diğer modellere kıyasla %0.8'lik bir başarı oranıyla öne çıktı. Ön işleme tekniklerinden köklerine ayırma yönteminin başarıyı artırdığı gözlemlendi. Word2vec kelime gömme yöntemi kullanan modellerde ise lojistik regresyon modeli diğerlerinden daha başarılı oldu (%0.72). Genel olarak ön işleme tekniklerinin kullanılması başarı oranlarında büyük bir değişikliğe sebebiyet vermedi. One hot encoder kelime gömme metodu kullanan modellerde Naive Bayes ve LR modelleri, farklı ön işleme yöntemleriyle benzer başarı oranları elde ederek %0.81 ve %0.80 oranlarında başarılı oldular. Bu çalışma, birçok doğal dil işleme tekniğinin bir arada kullanıldığı ve farklı kombinasyonlarının karşılaştırıldığı bir alana ışık tutmaktadır ve diğer çalışmalardan farklı olarak öne çıkmaktadır. Gelecekte yapılacak olan daha kapsamlı veri setleri ve farklı kelime gömme ve derin öğrenme tekniklerinin eklenmesiyle doğal dil işleme çalışmalarında bu bulguların rehberlik sağlayabileceğini ve geliştirilebileceğini düşünüyoruz.

**Kaynaklar**

[1] Çelik, Ö. & Koç, B. C. TF-IDF, Word2vec ve Fasttext Vektör Model Yöntemleri ile Türkçe Haber Metinlerinin Sınıflandırılması. Dokuz Eylül Üniversitesi Mühendislik Fakültesi Fen ve Mühendislik Dergisi 23, 121–127 (2021).

[2] Karcıoğlu, A. A. & Aydın, T. Sentiment analysis of Turkish and english twitter feeds using Word2Vec model in 2019 27th Signal Processing and Communications Applications Conference (SIU) (2019), 1–4.

[3] Lilleberg, J., Zhu, Y. & Zhang, Y. Support vector machines and word2vec for text classification with semantic features in 2015 IEEE 14th International Conference on Cognitive Informatics & Cognitive Computing (ICCI* CC) (2015), 136–140.

[4] Polpinij, J., Srikanjanapert, N. & Sopon, P. Word2Vec approach for sentiment classification relating to hotel reviews in Recent Advances in Information and Communication Technology 2017: Proceedings of the 13th International Conference on Computing and Information Technology (IC2IT) (2018), 308–316.

[5] Mikolov T, Chen K, Corrado G, Dean J. (2013), "Efficient estimation of word representations in vector space". Proceedings of Workshop at ICLR. Scottsdale, Arizona 2-4 Mayıs 2013.

[6] Long Ma and Yanqing Zhang. Using word2vec to process big text data. In 2015 IEEE International Conference on Big Data (Big Data), pages 2895–2897. IEEE, 2015.

[7] Tuncelli, O., & Özdemir, B. (2019). Turkish Stemmer for Python. GitHub. https://github.com/otuncelli/turkishstemmer-python.

[8] Cover, T., & Hart, P. (1967). Nearest neighbor pattern classification. IEEE transactions on information theory, 13(1), 21-27.

[9] Frank, E., & Bouckaert, R. R. (2006). Naive bayes for text classification with unbalanced classes. In Knowledge Discovery in Databases: PKDD 2006: 10th European Conference on Principles and Practice of Knowledge Discovery in Databases Berlin, Germany, September 18-22, 2006 Proceedings 10 (pp. 503-510). Springer Berlin Heidelberg.

[10] Shah, K., Patel, H., Sanghvi, D., & Shah, M. (2020). A comparative analysis of logistic regression, random forest and KNN models for the text classification. Augmented Human Research, 5(1), 12.

[11] Sjarif, N. N. A., Yahya, Y., Chuprat, S., & Azmi, N. H. F. M. (2020). Support vector machine algorithm for SMS spam classification in the telecommunication industry. Int. J. Adv. Sci. Eng. Inf. Technol, 10(2), 635-639.

[12] Luo, X. (2021). Efficient English text classification using selected machine learning techniques. Alexandria Engineering Journal, 60(3), 3401-3409.

[13] Wicaksono, F. A., & Romadhony, A. (2022). Sentiment analysis of university social media using support vector machine and logistic regression methods. Indonesia Journal on Computing (Indo-JC), 7(2), 15-24.

[14] Vadivu, G. Comparative Study of Logistic Regression and LSTM for Sentiment Classification Across Diverse Textual Dataset. parameters, 1, 4.

[15] Wang, W., & Sun, D. (2021). The improved AdaBoost algorithms for imbalanced data classification. Information Sciences, 563, 358-374.